\pdfoutput=1

\documentclass[11pt]{article}

\usepackage[]{acl}
\usepackage{times}
\usepackage{latexsym}

\usepackage[T1]{fontenc}
\usepackage[utf8]{inputenc}

\usepackage{microtype}
\usepackage{booktabs}
\usepackage{comment}
\usepackage{cleveref}
\usepackage{graphicx}

\title{HausaNLP at SemEval-2023 Task 10: Transfer Learning, Synthetic Data and Side-Information for Multi-Level Sexism Classification}

\author{Saminu Mohammad Aliyu$^{1+}$, Idris Abdulmumin$^{2+}$, Shamsuddeen Hassan Muhammad$^{1+}$, \\
{\bf Ibrahim Said Ahmad$^{1+}$, Saheed Abdullahi Salahudeen$^{3+}$, Aliyu Yusuf$^4$,}\\
{\bf Falalu Ibrahim Lawan$^{3+}$}\\
$^{1}$Bayero University, Kano, $^{2}$Ahmad Bello University, Zaria, $^{3}$Kaduna State University, \\
$^{4}$Universiti Teknologi PETRONAS,\\
$^{+}$HausaNLP\\
\texttt{smaliyu.cs@buk.edu.ng}}

\begin{document}
\maketitle
\begin{abstract}
We present the findings of our participation in the SemEval-2023 Task 10: Explainable Detection of Online Sexism (EDOS) task, a shared task on offensive language (sexism) detection on English Gab and Reddit dataset.
We investigated the effects of transferring two language models: XLM-T (sentiment classification) and HateBERT (same domain - Reddit) for multi-level classification into Sexist or not Sexist, and other subsequent sub-classifications of the sexist data. We also use synthetic classification of unlabelled dataset and intermediary class information to maximize the performance of our models. We submitted a system in Task A, and it ranked 49$^{th}$ with F1-score of 0.82.  This result showed to be competitive as it only under-performed the best system by 0.052\% F1-score.

\textbf{Content warning: All examples of sexism comments used are for illustrative purpose only.}
\end{abstract}
 
\section{Introduction}
Sexism is a form of written or verbal attack on women based on their gender and other identity \cite{kirk2023semeval}. In general, there is a global concern about the prevalence of hate on social media. Consequently, many studies have attempted to ensure that the social media remain safe for everyone \cite{JIANG2022100182}. In this paper, we described our experiments for the SemEval-2023 Task 10: Explainable Detection of Online Sexism (EDOS). The shared task was divided into 3 sub-tasks that are all aimed at predicting fine-grained information about the type of sexism that exists on social media sites of Gab and Reddit. More information is provided in \Cref{sec:task} and the task description paper \cite{kirk2023semeval}.

Our models were fine-tuned on two pre-trained language models, namely XLM-T \cite{barbieri-etal-2022-xlm} and HateBERT \cite{caselli2020hatebert}. While XLM-T is a pretrained language model trained on 198 million tweets, the HateBERT on the other hand is a pretrained language model based on BERT \cite{devlin2018bert} that was pretrained on a large scale English Reddit dataset for the detection of abusive language. We used only the dataset used by the task organizers in our experiment. See \citet{kirk2023semeval} for a description of the dataset. We made submission only to the Task A during the competition phase, and our system achieved a competitive performance of 0.82 F1-score. However, we also described our attempts at creating competitive models for Tasks B and C.

The main contributions of this paper are as follows: 
\begin{enumerate}
    \item We investigated the effectiveness of transferring two language models, namely, XLM-T  and HateBERT for binary sexism classification.  
    \item We explore the use of synthetic classification of unlabelled dataset and intermediary class information for maximizing the performances multi-level sexism classification models.
\end{enumerate}

\begin{figure*}
    \centering
    \includegraphics[width=0.85\textwidth]{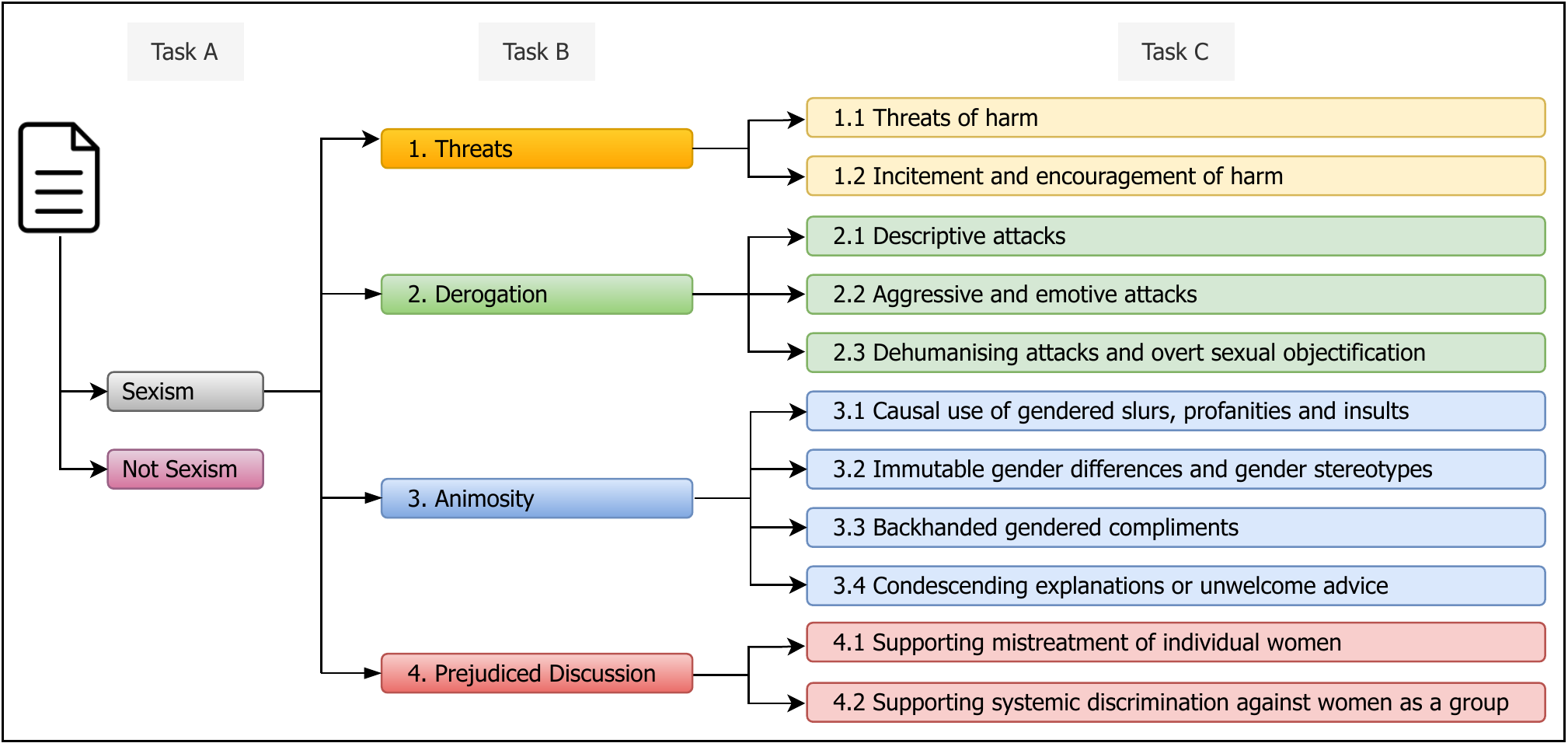}
    \caption{Different level of classification as provided in the shared task}
    \label{fig:my_label}
\end{figure*}

\section{Related Works}
There is an abundance of literature on the detection of hate speech online. However, only a fraction of such studies focused on sexism detection \cite{aliyu2022herdphobia}.

\citet{JIANG2022100182} proposed a Chinese dataset and lexicon for the detection of sexism on social media. The proposed dataset and lexicon were created from comments and posts collected on the Sina Weibo Chinese microbloging site and annotated into sexism or non-sexism. Those labelled as sexism were further annotated into four categories. Finally, each of these categories was labelled according to the target. Preliminary results show that context-based models outperform linguistics-based models. \citet{waseem2016hateful} created a dataset of 16,000 tweets, of which 3,383 tweets were sexist. Character n-gram feature was investigated alongside others such as gender and location. The best result was obtained with the combination of char n-gram and gender. 

At the IberEval 2018 task for the identification of misogyny from Spanish and English corpora, \citet{canos2018misogyny} used TF-IDF features vectors and SVM with a linear kernel to develop a system that classifies tweets as misogynistic or not misogynistic for subtask A, and the type and target of the misogynistic tweet for subtask B. The best result was obtained from the Spanish dataset for both tasks. Another study used TF-IDF, user-based, network-based, and text-based features together with classical and deep learning algorithms to automatically detect sexism in tweets. Logistic regression (LR), support vector machine (SVM) and random forest (RF) were used for the classical machine learning. Bidirectional long-short-term memory (Bi-LSTM) and multilingual bidirectional encoder representation from transformers (mBERT) from the deep learning algorithms. The mBERT with text features gave the best performance with an accuracy of 0.74. 

\cite{rodriguez2020automatic} conducted a review on racist and sexist hate speech detection, with special emphasis on datasets, features and approach used. According to their findings, \citet{waseem2016hateful} dataset was found to be the most used. Deep learning features perform best, and deep learning algorithms outperform the classical algorithms. \citet{istaiteh2020racist} reviewed studies on the detection of racist and sexist hate speech detection with special emphasis on datasets, features and approach used. The study concluded that the \citet{waseem2016hateful} dataset is the most widely used dataset and deep learning achieves better performance for the classification task.

\begin{table*}[t]
    \centering
    \begin{tabular}{lrrp{3.8in}}
    \toprule
        \textbf{Sentiment} & \textbf{\#Dev} & \textbf{\#Train} & \textbf{Example}\\
    \midrule
         Sexist & 340 & 3,544 & Western women are truly the most entitled people to have ever lived on this earth.\\
         
         Not Sexist & 1,060 & 11,056 & Our daughter was murdered, but at least we're not racists!\\
    \bottomrule
    \end{tabular}
    \caption{Dataset Description and Distribution of Sentiment Labels}
    \label{tab:my_datasets}
\end{table*}

\section{Task Overview}
\label{sec:task}
The SemEval-2023 subtasks aim to create  models for the detection of sexist post from Gab and Reddit. There are three subtask which include: Task A - a binary classification of statements as either sexist or not; Task B - classifies sexist statements into four groups namely, threats, derogation, animosity and prejudiced discussions; Task C - involves classification of the sexist statements into 11 fine-grained vectors. This is illustrated in \Cref{fig:my_label}.

\subsection{Train and development datasets}

It can be observed from \Cref{tab:my_datasets} that the dataset is imbalanced, with "Not Sexist" having more than 75\% in both train and dev splits. Both the train and dev datasets have similar distributions, with each having a minimum, maximum and average token counts of 1, 55 and about 23.5 respectively, based on space character (" ") tokenization.

\begin{figure}[!ht]
    \centering
    \includegraphics[width=\columnwidth]{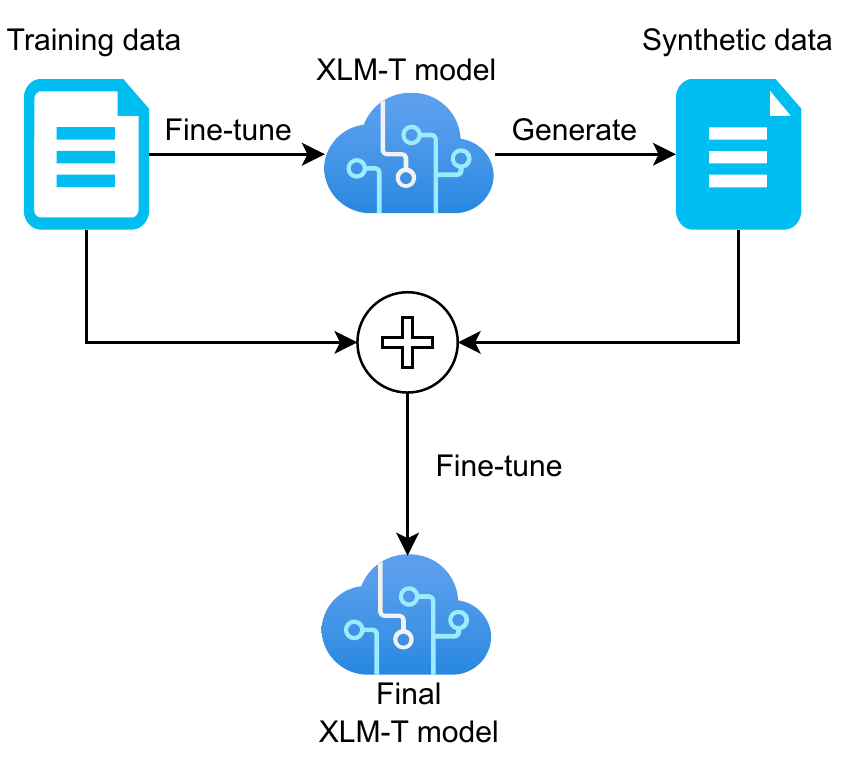}
    \caption{Task B. Using authentic and synthetic training datasets.}
    \label{fig:task_b}
\end{figure}

\begin{figure}[!ht]
    \centering
    \includegraphics[width=\columnwidth]{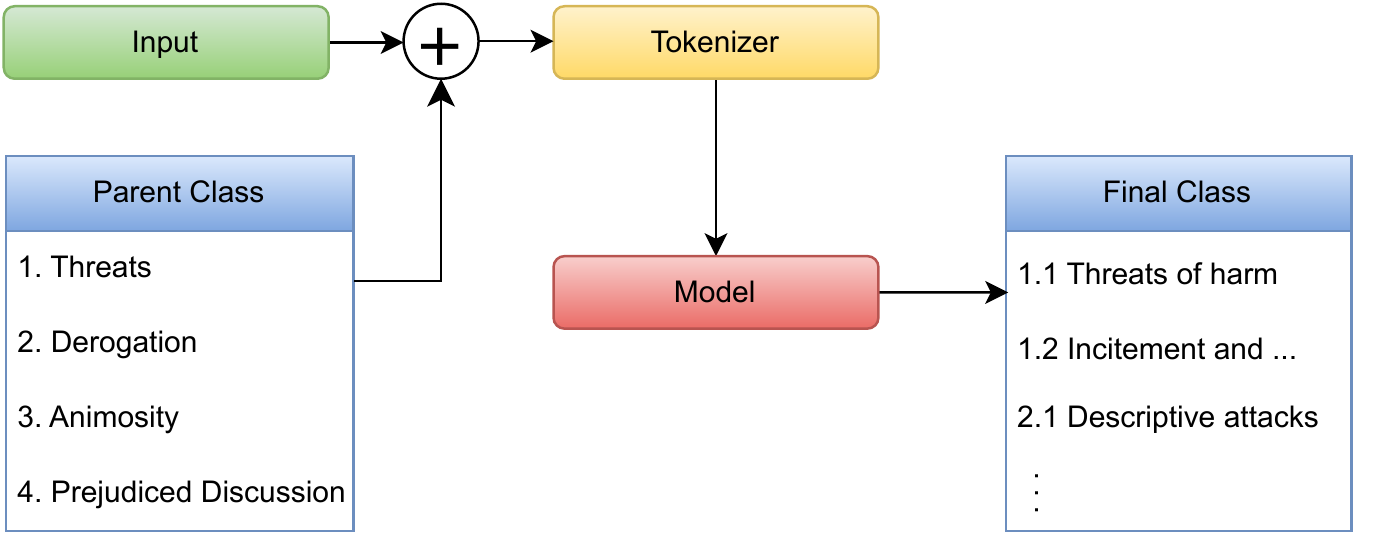}
    \caption{Task C. Each input sentence is paired with its parent class \textbf{["Threats", "Derogation", "Animosity", "Prejudiced Discussion"]} before tokenization.}
    \label{fig:task_c}
\end{figure}

\section{System Overview}
In this section, we will elaborate on the main methods for the binary sexism task.

\subsection{Pre-trained language models}
\label{plms}
For all the models, we fine-tuned two pre-trained language models (PLMs): XLM-T \cite{barbieri-etal-2022-xlm} and HateBERT \cite{caselli2020hatebert}.

\paragraph{XLM-T} This model was trained on 198 million tweets that were collected over a 2-year period starting from May 2018. The model was trained from a checkpoint of the base model, XLM-R \cite{conneau2019unsupervised}, until convergence\footnote{\url{https://github.com/cardiffnlp/xlm-t}}. This model was selected based on its excellent performance on sentiment analysis on social media datasets \cite{barbieri-etal-2022-xlm,aliyu2022herdphobia}.

\paragraph{HateBERT} This model is a BERT \cite{devlin2018bert} model that was retrained on a large scale English Reddit dataset\footnote{\url{https://huggingface.co/GroNLP/hateBERT}} \cite{caselli2020hatebert} for the detection of abusive language. The dataset consisted of banned abusive, hateful and offensive words. We used this model because sexism is a form of offensive and hateful language, but also because the re-training dataset was from the same social platform as the provided training dataset.

\subsection{Training Strategies}

For Task A - the binary classification task, two models were trained by fine-tuning the pre-trained language models (PLM) mentioned above. We used the best model in this task to select sentences that are potentially sexist from the provided unlabeled. In Task B, we first used the training data provided to fine-tune the XLM-T PLM. Subsequently, we used the fine-tuned model to generate the automatic classification of the potentially sexist sentences that were generated using the model in Task A. We then mixed this bigger synthetic dataset with the authentic data to train another classification model. This is illustrated in \Cref{fig:task_b}.

Finally, for Task C, we leveraged the parent classification of the input sentence as provided for Task B to help the model narrow the expected final sexism classification, as illustrated in \Cref{fig:task_c}. We utilized this information because it is provided in the shared task. For real-world application, we understand that this information may not be available. For this, we anticipate using a model to predict the parent classes, before using the synthetic data as the side information.



\section{Experimental Setup and Results}
\subsection{Dataset}
For this task, we only used the dataset provided by the organizers. We used a subset of training data (10\%) to develop the systems and maintained the same set during the competition phase. We added the released development data to the other 90\% of the training dataset to create a new train set, while maintaining the original development split. Some useful statistics of these datasets are provided in \Cref{tab:train_dev_test}. For the test set, a total of 4,000 provided for the competition, and we used them as they are to evaluate the performances of the various models.

\begin{table}[]
    \centering
    \begin{tabular}{lrr}
    \toprule
        \textbf{data} & \textbf{\# sentences} & \textbf{\# tokens} \\
    \midrule
         train & 14,600 & 407,857 \\
         dev & 1,400 & 39,326 \\
         test & 4,000 & 110,282 \\
    \bottomrule
    \end{tabular}
    \caption{Data split}
    \label{tab:train_dev_test}
\end{table}

\subsection{Models}
For the models, we used the publicly available checkpoints in Huggingface,\footnote{\url{https://huggingface.co/GroNLP/hateBERT}}$^,$\footnote{\url{https://huggingface.co/cardiffnlp/twitter-xlm-roberta-base-sentiment}} training them for 20 epochs using a training batch size of 32 and a maximum sequence length of 128. We use the code\footnote{\url{https://github.com/IyanuSh/YOSM}} and default model hyper-parameters as provided in \citet{shode_africanlp} to train the models. The configuration uses Adam \cite{KingmaB14} for optimization, an initial learning rate of $5e-5$, and $1e-8$ epsilon for the Adam optimizer.

\subsection{Results}
Our experimental results are presented in \Cref{tab:results}. From these results, we found that the XLM-T model achieved the best performance on the binary classification task. Even though the HateBERT model was fine-tuned on the abusive and banned data from Reddit (same domain as some of the training and evaluation data), the model was not able to outperform the XLM-T. We will conduct more in-depth experiments to determine the actual reason for this anomaly.

\begin{table}[]
    \centering
    \begin{tabular}{llr}
    \toprule
       Task & Method & F1 \\
    \midrule
        A & XLM-T & 0.8228 \\ 
        A & HateBERT & 0.8172 \\ \hline
        B & XLM-T & 0.5981 \\
        B & XLM-T+synth & 0.6012 \\ \hline
        C & XLM-T & 0.3565 \\
        C & XLM-T+parent class & 0.4151 \\
    \bottomrule
    \end{tabular}
    \caption{Result of the different tasks.}
    \label{tab:results}
\end{table}

For Task B, a slight performance was realized after adding the synthetic data, even though the quality of such data is less than the labelled data. This further reinforces the fact that more data is more often than not beneficial to neural models.

For Task C, we recorded a substantial improvement in the performance of the model after supplementing the side information (the parent class) to influence its prediction. Rather than just passing a sentence and expecting the model to predict over 11 classes, the model performed better when its parent class is known to it at the tokenization stage. We anticipate a slight drop in performance if the parent class information is synthetic rather than, in this case, authentic. However, we cannot substantiate the truthfulness or not of this claim, or the extent of the effect, without conducting further experiments.

\subsection{Competition rank}
We submitted only the XLM-T model in Task A to the competition and although our model ranked 49$^{th}$, the best model in the competition track only outperformed our model by 0.052\%, as indicated in \Cref{tab:comp_results}.

\begin{table}[]
    \centering
    \begin{tabular}{llr}
    \toprule
        \textbf{\#} & Team & F1 \\
    \midrule
        1 & PingAnLifeInsurance & 0.8746 \\
        2 & stce & 0.8740 \\
        3 & FiRC-NLP & 0.8740 \\
        4 & PALI & 0.8717 \\
        5 & GZHU / UCAS-IIE & 0.8692 \\
        \vdots & \vdots & \vdots \\
        49 & HausaNLP & 0.8228 \\
        \vdots & \vdots & \vdots \\
        84 & NLP\_CHRISTINE & 0.5029 \\
    \bottomrule
    \end{tabular}
    \caption{Task A official ranking}
    \label{tab:comp_results}
\end{table}

\section{Conclusion}
In this system description paper, we describe our submission for the subtask A binary classification of comments into sexist and not sexist submitted to the SemEval-2023 Task 10 - Explainable Detection of Online Sexism (EDOS). We implemented pretrained models using XLM-T and HateBERT on the English Language Twitter. Our model achieved competitive result of 82\% using F1 score, slightly below the leader with 87\%. However, this performance is not exhaustive as we observe great imbalance in the distribution of the dataset and this could have influenced the results. Furthermore, we described our attempts at building competitive models for Tasks B and C (which we did not submit to the competition). We utilised synthetic data and parent class information to improve the performances of the two models in the respective tasks, and some improvements were observed. Finally, we intend to improve the performances of these models by targeting the data imbalance constraint using data augmentation strategies.

\bibliography{anthology,custom}
\bibliographystyle{acl_natbib}

\appendix

\label{sec:appendix}


\end{document}